\documentclass[twocolumn]{article}

\usepackage[width=17cm,height=22cm]{geometry}
\usepackage[english]{babel}
\usepackage[utf8]{inputenc}
\usepackage{fancyvrb}
\usepackage{authblk}
\usepackage{graphicx}
\usepackage{hyperref}
\usepackage{amsmath, amssymb}

\usepackage{xcolor}
\usepackage{placeins}
\usepackage{enumitem}

\newcommand{\corGG}[2]{{\color{red}#2}}

\title{Dilated Spatial Generative Adversarial Networks for Ergodic Image Generation}
\author[1]{Cyprien Ruffino}
\author[1]{Romain Hérault}
\author[2]{Eric Laloy}
\author[1]{Gilles Gasso}
\affil[1]{Normandie Univ, UNIROUEN, UNIHAVRE, INSA Rouen, LITIS, 76~000 Rouen, France\footnote{This research was supported by the CNRS PEPS I3A  REGGAN project and the ANR-16-CE23-0006 grant \emph{Deep in France}}}
\affil[2]{Belgian Nuclear Research, Institute Environment, Health and Safety, Boeretang 200 - BE-2400 Mol, Belgium}

\begin{document}
\maketitle

\begin{abstract}
Generative models have recently received renewed attention as a result of adversarial learning. Generative adversarial networks consist of samples generation model and a discrimination model able to distinguish between genuine and synthetic samples. In combination with convolutional (for the discriminator) and deconvolutional (for the generator) layers, they are particularly suitable for image generation, especially of natural scenes. However, the presence of fully connected layers adds global dependencies in the generated images. This may lead to high and global variations in the generated sample for small local variations in the  input noise. In this work we propose to use architectures based on fully convolutional networks (including among others dilated layers), architectures specifically designed to generate globally ergodic images, that is images without global dependencies. Conducted experiments reveal that these architectures are well suited for generating natural textures such as geologic structures.
\end{abstract}

\section{Introduction}

Using Deep Generative models to generate images of the subsurface rock structure has been proposed by Laloy et al. \cite{Laloy2017_1,laloy2018training}. In this study we improve upon the work by \cite{laloy2018training} who generated geologic images using fully convolutional Generative Adversarial Networks.

\begin{figure}[!ht]
    \centering
    \includegraphics[width=0.45\columnwidth]{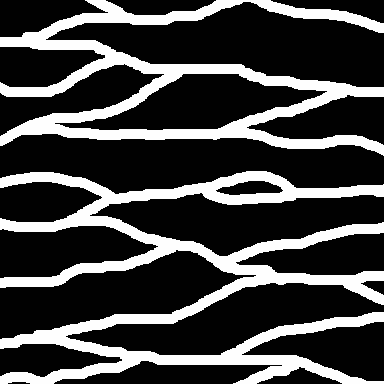}
    \includegraphics[width=0.45\columnwidth]{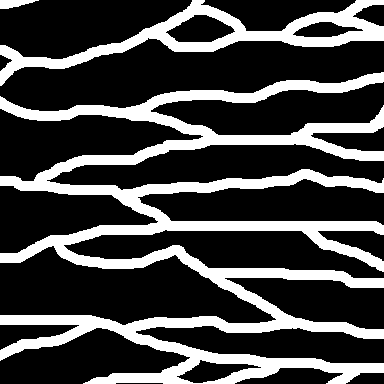}
  	\caption{Classical 2D toy model of the subsurface used in the geosciences. The model consists of sand channels (white) in a clay matrix (black). The aim of the presented work is to synthesize such images.}
    \label{figure:groundtruth}
\end{figure}

    Generative Adversarial Networks \cite{Goodfellow2014} have been recently highlighted for their ability to generate high quality images. Moreover, the generation process is quite easy to do and fast once the networks are trained. Indeed, it simply consists of sampling a noise in the input distribution, which is usually Gaussian or uniform, and computing a generator forward pass. 
    
    However, and even if the subsurface models are analogous to images, Generative Adversarial Networks show some limitations as they do not preserve the global ergodicity of the generated data. An image is globally ergodic if each sub-sample of this image shares the same properties as the original image or in other words, there is no global dependencies in the image. In several works including \cite{Laloy2017_1, laloy2018training}, optimization is done on the output space of the generator using methods such as Bayesian inversion, which are made significantly easier when no global dependencies are present in the generated data.
    
    In this paper, we propose a new architecture based on Spatial Generative Adversarial Networks \cite{Jetchev} with added dilated convolutions for globally ergodic data generation. Dilated convolutions are a variant of convolutional layers in which the filters are sparse and of adjustable size. This allows to have filters with a large receptive field without having to change the size of the data via pooling or striding. We show that this method perform better than existing ones producing, notably, less noisy and blurry results. 
    
    The remainder of the paper is as follows: section \ref{section:relatedwork} describes the related works and formally present the building blocks of our architecture. Our approach is then explained in section \ref{section:approach}, and experiments are detailed in section \ref{section:experiments}.

\section{Related work}
\label{section:relatedwork}

    In this section, we introduce the Generative Adversarial Networks framework. We then present its fully-convolutional variation, Spatial Generative Adversarial Networks, which is more adapted to the task of ergodic image generation.
    
	\subsection{Generative Adversarial Networks}

		Generative Adversarial Networks \cite{Goodfellow2014} are generative models learned in an unsupervised way. Given training samples, instead of learning their underlying density function, GAN attempt to learn how to generate new samples by passing a random variable with some specified distribution through a non-linear function, typically a deep network. The generating network is devised such that the distribution of the generated samples is aligned with the unknown real distribution. GAN have gained an increasing popularity as they are able to produce high-quality images, especially natural scenes \cite{salimans16}.

        The training process of classical GANs consists in a zero-sum game between a generation model $G$ and a discrimination function $D$, in which the generator learns to produce new synthetic data and the discriminator learns to distinguish real examples from generated ones. Both $G$ and $D$ are usually some flexible functions such as deep neural networks. More formally, training GANs is equivalent to solving the following saddle-point problem, where $x$ represents a real sample drawn from an unknown distribution $P_r$ and $z$ is a noise input, sampled from a known probability distribution $P_z$:
		\begin{align}
        \label{eq:ganclassic}
			\min_G \max_D  \mathop{\mathbb{E}}_{x\sim P_r} \log D(x) 
            + \mathop{\mathbb{E}}_{z\sim P_z} \log (1-D(G(z)))
		\end{align}

		Goodfellow et al.  \cite{Goodfellow2014} established that solving the minimax problem amounts to minimize Jensen-Shannon divergence between $P_r$ and $P_z$. In practice, a slightly different formulation is considered to avoid vanishing gradient issues: the generator $G$ maximizes  $L_G$ while the discriminator $D$ minimizes $L_D$ (both objective functions are stated in equations \ref{eq:LG} and \ref{eq:LD}).
		\begin{align}
        \label{eq:LG}
			L_{G} &= - \mathop{\mathbb{E}}_{z\sim P_z} [\log D(G(z))]\\
        \label{eq:LD}
			L_{D} &=  \mathop{\mathbb{E}}_{x\sim P_r} [\log D(x)] + \mathop{\mathbb{E}}_{z\sim P_z} [1 - \log D(G(z))]
		\end{align}
We apply this strategy to train our proposed model. Finally, notice that a summarized overview of unconditional GAN and their evaluation are described in \cite{lucic2017gans}

	\subsection{Spatial GAN}

		Spatial Generative Adversarial Networks (SGANs) \cite{Jetchev} represent a sub-category of GANs in which both the generator and the discriminator models are fully-convolutional networks. They are based on the previous Deep Convolutional Generative Adversarial Networks \cite{Radford2015} architecture, which are more adapted to the task of image generation.

	As a consequence of using fully-convolutional networks, the output of the discriminator may not be single scalar, but rather a matrix of $n\times m$ values in $[0, 1]$. The objective function (\ref{eq:ganclassic}) is adapted accordingly for SGAN by averaging over the discriminator's output:
		\begin{align}
			\min_G \max_D  & \mathop{\mathbb{E}}_{x\sim P_r} \frac{1}{m}\frac{1}{n}\sum_{i=1}^{m}\sum_{j=1}^{n} \log D(x)_{i, j} \nonumber \\
            &+ \mathop{\mathbb{E}}_{z\sim P_z} \frac{1}{m}\frac{1}{n}\sum_{i=1}^{m}\sum_{j=1}^{n}  \log (1-D(G(z))_{i, j})
		\end{align}
The term $D(u)_{i, j}$ stands for the $(i,j)$ entry of the discriminator output $D(u)$. SGAN brings several interesting properties.  First, as both networks are fully-convolutional, any decision taken in the discriminator and the generator is, for a given location $(i,j)$, solely based on a local context as the amount of information both $G$ and $D$ have is limited by the receptive field of the filters of their deep networks. This implies that the networks are unable to model global dependencies in the data, meaning that global ergodicity tend to be preserved in the generated images.

Then, as both networks are fully-convolutional, there is no restriction on the input size, in both the training process and the generation phase.

Finally, because fully-connected layers usually contain the majority of the network's weights, SGAN tends to have far less parameters than \corGG{more}{} classical GANs.

\subsection{Application to geologic structure generation}

\begin{figure}[!ht]
    \centering
    \includegraphics[width=0.45\columnwidth]{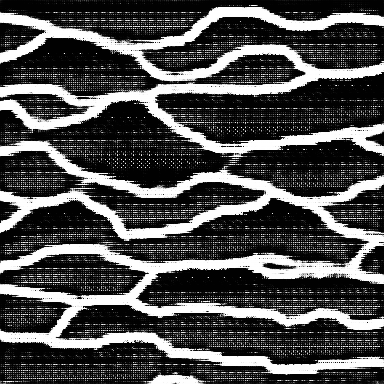}
    \includegraphics[width=0.45\columnwidth]{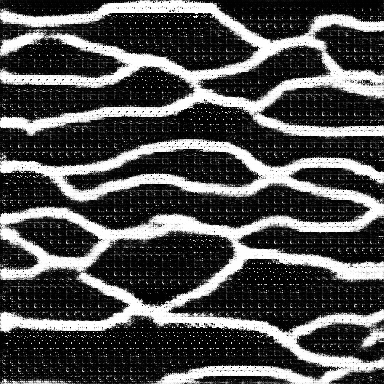}
  	\caption{Samples generated with the SGAN approach. We can see that noise is present in these images, and some channels are blurry, especially near intersections.}
    \label{figure:blurry}
\end{figure}

In \cite{laloy2018training}, authors showed that Spatial Generative Adversarial Networks works well when the generated data is locally structured, yet globally ergodic, making them viable for geostatistical simulation (e.g, simulation of subsurface spatial structures). However, the generated samples tended to be noisy, blurry or to have visual artifacts (see Figure \ref{figure:blurry}). To overcome this problem, ad-hoc solutions like median filtering or thresholding the generated images are set up at a post-processing stage \cite{laloy2018training}.

\section{Proposed approach}
\label{section:approach}

The purpose of the paper is to enhance the quality of generated images in order to get rid of these ad-hoc methods. To attain this goal, we propose to use dilated convolutions rather than classical ones. Dilated convolutions allow to learn filters with large receptive fields, hence are more able to handle global ergodicity of ground simulated images. Before delving into the details of proposed architecture we briefly introduce dilated convolutions.

\subsection{Dilated convolutions}

\begin{figure}[!h]
\includegraphics[scale=0.465]{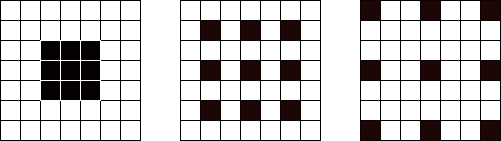}
\caption{Dilated filters with dilation rate of 1, 2, 3}
\label{fig:dilatation}
\end{figure}

    	Dilated convolutions \cite{Yu} (or ``A trous'' convolutions) are convolutions in which the size of the filters receptive fields are artificially increased, without increasing the number of parameters, by using sparse filters\ref{fig:dilatation}.
        This method was first introduced in the \textit{algorithme à trous} \cite{Holschneider1988} for wavelet decomposition, and was recently adapted to convolutional neural networks.






        Dilated convolutions allows to control the size of the receptive fields of a filter without changing the dimension of the data via pooling or striding and without increasing the number of parameters. This property is especially useful in tasks where the precise pixel position is important, like image segmentation.
For this particular task, dilated convolutions have shown good performance with both classical deep convolutional networks \cite{Chen} and fully-convolutional architectures \cite{Hamaguchi2017}.

\subsection{Dilated SGAN for geologic structure generation}
\begin{figure*}
  \includegraphics[width=\textwidth,height=2.9cm]{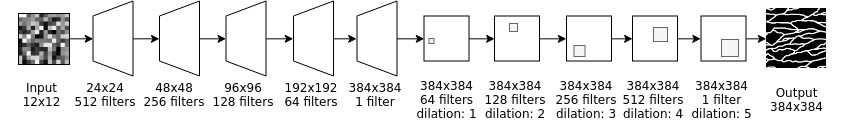}
  \caption{The architecture of our generator network. The first five layers are deconvolutional layers with strides of $1/2$, and are followed by five dilated convolutional layers with dilation rates of 1, 2, 3, 4 and 5.}
  \label{figure:architecture}
\end{figure*}

Our main contribution in this work is the use of an architecture based on the original SGAN design, in which we introduced dilated convolutions at the higher layers of the generator. The generator we design consists of 2 parts:
\begin{itemize}[noitemsep,topsep=0pt]
\item a series of deconvolution layers, usually with striding, in order to scale up the input noise to the right output size, as in standard SGAN,
\item a new sequence of dilated convolution layers.
\end{itemize}
Figure \ref{figure:architecture} shows the principle of proposed generator. Implementation details are provided in the next section.

No modification is made to the discriminator compared to SGAN/DCGAN. Hence the discriminator is a fully-convolutional network that consists of several convolutional layers with striding. As for SGAN/DCGAN, each output pixel of the discriminator indicates if the part of the input at its receptive field is true or generated.

\section{Experiments}
	We run experiments on generating images of geologic structures using our dilated convolutions-based SGAN. We show that substantial benefits are obtained in terms of visual quality of produced images and numerical evaluation metrics.

\label{section:experiments}
	\subsection{Network architecture}

We base our adversarial generation model on the SGAN architecture \cite{Jetchev}. Implementation details are as followed:

The generator is constituted of five deconvolutional layers with strides of $1/2$ and $5\times 5$ filters. From \cite{Jetchev}, we added five dilated convolutions with dilation rates increasing from one to five. All layers in the generator have ReLU activations (except for the last layer which uses tanh). Moreover, the last layer has a number of filters equal to the number of channels of the output image. The number of filters in each layer in detailed in Figure \ref{figure:architecture}.
Batch normalization \cite{Szegedy2015} is added between all layers except before the last deconvolution and before the last dilated convolution. 

The discriminator is a five layer convolutional network with strides of $2$ and $9\times 9$ filters. The last convolutional layer only has one filter with a sigmoid activation function.

	\subsection{Experimental setup}

    	In our experiments, we used the same dataset as \cite{laloy2018training}, which consists of images of size $384\times 384$ sampled from a $2500\times 2500$ classical toy model of a complex subsurface binary domain (see Figure \ref{figure:groundtruth}). As this data is globally ergodic and we only try to learn its local structure, each sample is meaningful and informative (its provides a reduced view of a larger ground structure). The same process was adopted by Jetchev et al. \cite{Jetchev} who learn to generate textures from a large texture image by sampling smaller patches in it. An example of our training images is shown in Figure \ref{figure:groundtruth}. All training images pixels have their values set  between -1 and 1, as recommended in \cite{Radford2015}.
        
        The input noise of the generator  belongs to $[-1, \quad 1]^{12\times 12}$ and is sampled from a uniform distribution. The input size $12\times 12$ is chosen in order that deconvolution layers lead to an output image of size $384\times 384$.

		We used the Adam \cite{Kingma} optimizer with a learning rate of $5\times 10^{-4}$ and a $\beta_1$ value of $0.5$ for both the generator and the discriminator as in \cite{Jetchev}. L2 regularization with $\lambda=10^{-5}$ is added to every layer. We trained the network for 100 epochs, with a minibatch size of 8. In each epoch, we train both the generator and the discriminator on 100 minibatches.

\begin{figure}[!ht]
    \centering
    \includegraphics[width=0.45\columnwidth]{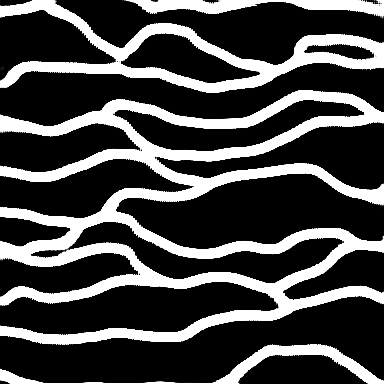}
    \includegraphics[width=0.45\columnwidth]{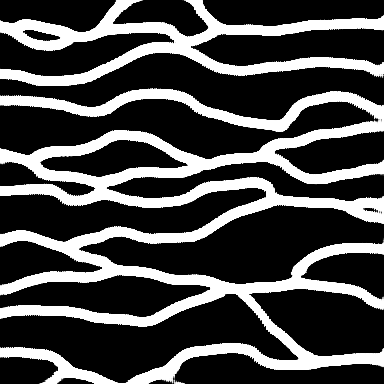}
    \includegraphics[width=0.45\columnwidth]{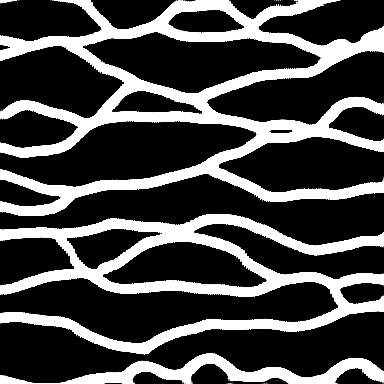}
    \includegraphics[width=0.45\columnwidth]{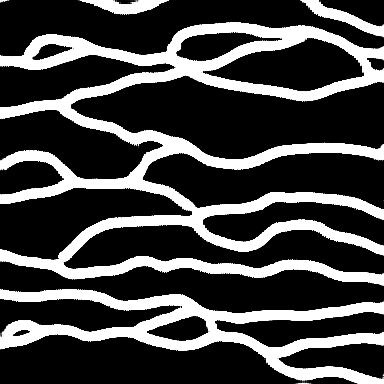}
  	
  	\caption{Four samples generated with our approach.}
    \label{figure:dilationapproach}
\end{figure}

Our method is implemented using Keras and TensorFlow\cite{Abadi}. The training process took from three to four hours on a NVidia GeForce GTX 1080Ti. Some generated samples can be seen at Figure~\ref{figure:dilationapproach}        

    \subsection{Evaluation}

To evaluate our results, we used a domain-specific metric, namely the connectivity function (also called cluster function) of the image \cite{Torquato1988} using the code by \cite{lemmens2017}, to estimate the quality of the generated samples from the application point of view. This metric is an effective structural descriptor and is commonly used in present geostatistical applications. 
        
To assess the visual quality of our samples, we use more classical image evaluation metrics: the total variation norm \cite{Rudin1992} and the mean $\chi^2$ distance between histograms of features  taken from real data and generated samples. Two kind of histograms of visual descriptors have been computed: local binary patterns (LBP) \cite{Pietikainen2011} and histograms of oriented gradients (HOG) \cite{Dalal}. These descriptors are commonly used in texture classification.

The \textbf{connectivity function} \cite{Torquato1988,lemmens2017} is the probability that a continuous pixel path exists between two pixels of the same class (or facies) separated with a given distance (called lag), in a given direction. In other words, it is the probability for two pixels separated with a given distance to be in the same cluster. In our application domain, clusters can be thought as the same sand channel or clay matrix zone.
In this work, we compute these functions along the X and Y axis. 

\textbf{Total variation norm} (TV) \cite{Rudin1992} is a good indicator of the noisiness of our data, as noisy signals tend to have a higher total variation. Total variation measure is frequently used in denoising tasks, as a criterion to minimize or as a regularization. We compute both the isotropic total variation $TV_i$ and its anisotropic variant $TV_a$ with :
        \begin{align}
        	TV_i(y) &= \sum_{i,j} \sqrt{|y_{i+1, j} - y_{i, j}|^2 + |y_{i, j+1} - y_{i, j}|^2} \\
        	TV_a(y) &= \sum_{i,j} |y_{i+1, j} - y_{i, j}| + |y_{i, j+1} - y_{i, j}|
        \end{align}

\textbf{Local binary patterns} (LPB) \cite{Pietikainen2011} are a visual descriptor that is frequently used in texture classification tasks. This descriptor makes sense in our use-case since our data is globally ergodic. It consists in extracting local patterns comparing the light-level of a pixel with its neighbors by dividing an image into cells of radius $R$ and computing a histogram of binary light levels (1 if the pixel is brighter than the center of the cell, 0 otherwise). 

\textbf{Histograms of oriented gradient} (HOG) \cite{Dalal} are a feature descriptor that tend to highlight contours in images, and are often used for image classification. They are obtained by splitting the image into cells and computing the gradient in these cells, for example with a derivative filter, then computing the histograms of these gradients.

	\subsection{Results}
	
\begin{figure}[!h]
	\begin{center}
		\includegraphics[width=0.48\columnwidth]{gt}\\
		\includegraphics[width=0.48\columnwidth]{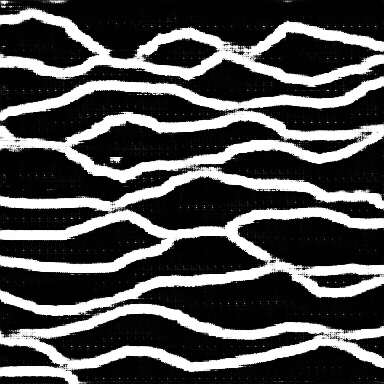}
		\includegraphics[width=0.48\columnwidth]{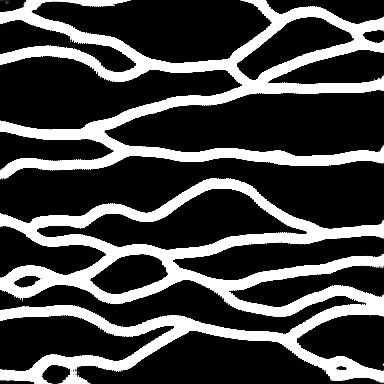}
	\end{center}
	\caption{Top: An image sampled from the real dataset; Bottom left: an image generated with the SGAN approach; Bottom right: an image generated with our approach.
	}
	\label{figure:images}	
\end{figure}

We compare the results of our approach with the SGAN approach presented in \cite{laloy2018training} using the aforementioned metrics on 100 generated samples by each approach.
At first, we are able to generate samples that are visually much less noisy and blurry.
Figures \ref{figure:dilationapproach} and \ref{figure:images} illustrate this fact.

        \begin{table}[!h]
	\centering
	
	\begin{tabular}{|l||c||c|c|}
		\hline
		& Real & SGAN & Our approach \\
		\hline
		$TV_i$& $\mathbf{5.37\times 10^{-2}}$ & $7.41\times 10^{-2}$ & $\mathbf{6.07\times 10^{-2}}$ \\
		\hline
		$TV_a$& $\mathbf{5.72\times 10^{-2}}$ & $8.02\times 10^{-2}$ &$\mathbf{6.53\times 10^{-2}}$\\
		\hline
	\end{tabular}
	\caption{Total variation norm on real and synthetic samples.}
	\label{table:tv}
\end{table}

\begin{table}[!h]
	\centering
	
	\begin{tabular}{|l||c|c|c|}
		\hline
		& SGAN & Our approach\\
		\hline
		LBP $R=1$ & $10.13$ & $\mathbf{2.39}$ \\
		\hline
		LBP $R=2$ & $24.26$ & $\mathbf{2.33}$ \\
		\hline
		HOG & $5.79\times 10^{-4}$ & $\mathbf{2.37\times 10^{-4}}$ \\
		\hline
	\end{tabular}
	\caption{Mean $\chi^2$ distance between histograms of real and synthetic data for LBP (with a radius $R$) and HOG features.}
	\label{table:features}
\end{table}

\begin{figure*}
        \centering
        \large{Laloy et al.}
        \includegraphics[scale=0.55]{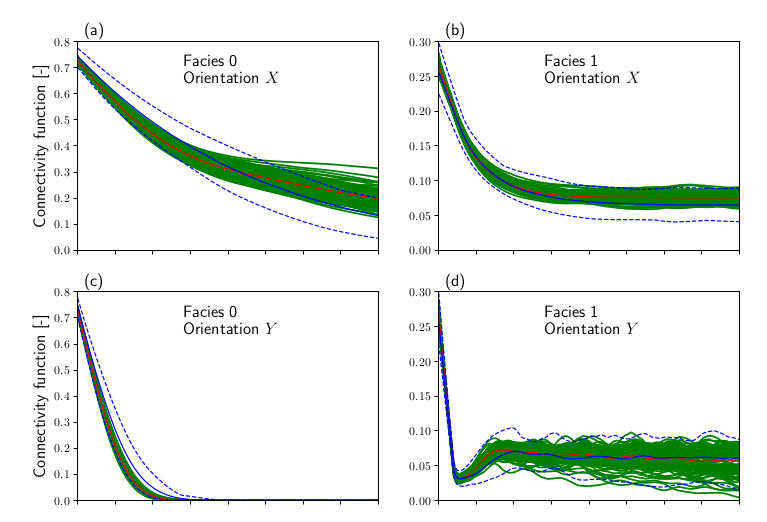}\\
        \large{Our approach}
        \includegraphics[scale=0.55]{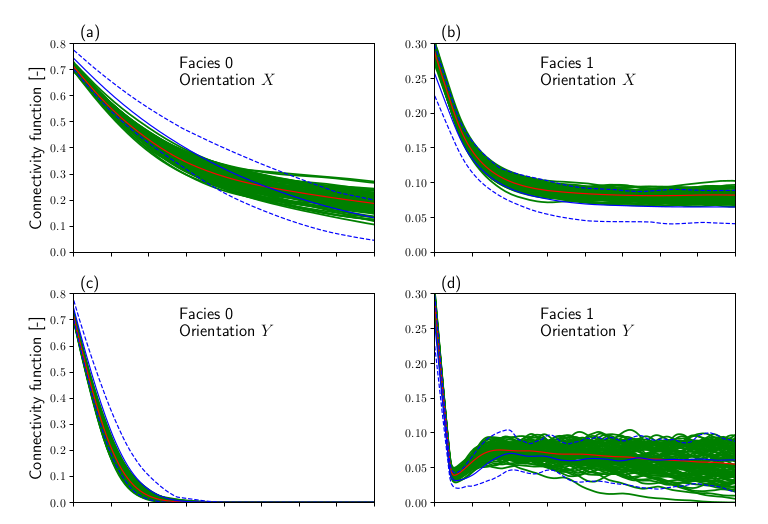}
        \caption{Connectivity function curves for both the SGAN approach  and our method. The green curves are the connectivity functions for every one of the 100 synthetic samples on which both the methods were evaluated on. The blue dashed curves indicates the maximum and the minimum values obtained on the real samples, and the red curve is the mean connectivity function the synthetic samples. Our goal is to have the most similar connectivity functions possible for real and synthetic images. We can see that our approach performs similarly as the SGAN approach used in Laloy et al. \cite{laloy2018training}.}
        \label{figure:connectivity}	
	\end{figure*}

Moreover, our approach presents similar connectivity functions as the SGAN approach (see Figure \ref{figure:connectivity}), meaning that enhancing the visual quality of the samples has no negative impact on the geostatistical quality of generated samples using our dilated convolutions SGAN.

Nevertheless, the total variation norm of the generated samples is lower with our approach, as shown in Table \ref{table:tv}, and closer to the the total variation of the real data. This is coherent with the fact that no visible noise is present in the images generated by our method.

Finally, it also performs better when we compute the mean $\chi^2$ distance between the real and generated samples for both LBP (computed with 8 neighbors and radii of 1 and 2) and HOG (Table \ref{table:features}).

\section{Conclusion}
	In this paper, we have presented an architecture for globally ergodic data generation. We have shown that our method produces samples that are sharper and less noisy than the previous approaches, removing the need for ad-hoc solutions like median filters, without altering their quality.
    
    We plan to extend our method to 3D image generation, as real-world geologic structures are essentially 3D.
	it is also our intention to make the method able to honor spatial constraints such as exact pixel or voxel values, or mean value over a given block. Such capability would be very useful for geostatisical applications.

{
\footnotesize
\bibliography{cap2018}
}

\FloatBarrier

\end{document}